\documentclass[conference]{IEEEtran}
\IEEEoverridecommandlockouts
\usepackage{cite}
\usepackage{graphicx}
\usepackage{textcomp}
\usepackage{xcolor}
\usepackage{fancyvrb}
\usepackage{listings}
\IEEEoverridecommandlockouts
\usepackage{cite}
\usepackage{algorithmic}
\usepackage{graphicx}
\usepackage{textcomp}
\usepackage{xcolor}
\usepackage{fancyhdr}
\usepackage{lastpage}
\def\BibTeX{{\rm B\kern-.05em{\sc i\kern-.025em b}\kern-.08em
    T\kern-.1667em\lower.7ex\hbox{E}\kern-.125emX}}

\newcommand{\texttttoken}[1]{\texttt{\detokenize{#1}}}

\def\BibTeX{{\rm B\kern-.05em{\sc i\kern-.025em b}\kern-.08em
T\kern-.1667em\lower.7ex\hbox{E}\kern-.125emX}}
\begin{document}

\title{NeuSymMS: A Hybrid Neuro-Symbolic \\ Memory System for LLM Agents}


\author{
\IEEEauthorblockN{Mujahid Sultan\IEEEauthorrefmark{1}\IEEEauthorrefmark{2}\thanks{M.~Sultan's prior peer-reviewed work includes DOI-registered publications in AI, enterprise architecture, and genetics; see ORCID 0000-0001-6721-4044.}, 
Sri Thuraisamy\IEEEauthorrefmark{1}, 
Daya Rajaratnam\IEEEauthorrefmark{1}}
\IEEEauthorblockA{\IEEEauthorrefmark{1}\textit{iVedha Corporation, Toronto, Ontario, Canada}\\
\IEEEauthorblockA{\IEEEauthorrefmark{2}\textit{MLSoft Inc., Toronto, Ontario, Canada}}\\
\texttt{mujahid.sultan@mlsoft.ai}, \texttt{sri@ivedha.com}, \texttt{daya@ivedha.com}}
}

\markboth{IEEE Transactions on X,~Vol.~X, No.~X, Month~Year}%
{Author \MakeLowercase{\textit{et al.}}: NeuSymMS: A Hybrid Neuro-Symbolic Memory System for LLM Agents}

\maketitle

\begin{abstract}
We present NeuSymMS, an adaptive memory system that enables large language model (LLM) agents to learn, remember, and reason about users across sessions via a hybrid neuro-symbolic architecture. NeuSymMS couples neural fact extraction from unstructured dialogue using LLMs and a CLIPS-based expert system that classifies, deduplicates, and reconciles facts under explicit lifecycle rules. The system represents knowledge as subject-relation-value triples stored in relational database management system. It supports user/agents/agent-to-agent scoping, and implements a dual-horizon (short-term and long-term) memory model. IT leverages access-based promotion and time-based pruning of the memory on both horizpons. NeuSymMS maintains continuity of memory while avoiding context-window bloat and cross-entity contamination. We argue that this architecture offers a practical path to trustworthy, auditable memory for production agentic systems and discuss its novelty relative to log retrieval, summarization, and key-value approaches.
\end{abstract}


\section{Introduction}

\IEEEPARstart{L}{arge} language model (LLM)-based assistants behave as persistent agents rather than single-turn chatbots, yet most deployed systems still treat conversation history as ephemeral context. Persistent memory for LLM agents has largely emerged from three families of techniques: conversation-log retrieval, LLM-based summarization, and key-value or knowledge-graph storage. These approaches are noisy, expensive, and ill-suited to handling contradictions or temporal change in user state \cite{liu2023lost,lewis2020retrieval,zhang2020pegasus}. As a result, many platforms either discard context after a session or naively append histories, leading to context-window bloat and brittle behavior when users change jobs, locations, preferences, or constraints \cite{openai2023gpt4,beltagy2020longformer}.

\textbf{Conversation-log retrieval :} The LoCoMo benchmark by Snap Research (ACL 2024) demonstrated that while long-context LLMs and RAG improve memory capabilities by 12--20\%, they still lag significantly behind human performance, especially in temporal reasoning (by 41\%) \cite{ahn2024locomo}.

\textbf{LLM-based summarization :} Wang et al. (Neurocomputing 2025) proposed recursive summarization that generates real-time summaries to enhance long-term dialogue memory, allowing LLMs to manage exceptionally long contexts across multiple sessions without increasing maximum length settings \cite{wang2025recursively}.

\textbf{Key-value or knowledge-graph storage :} REMem (ICLR 2026) introduced a hybrid memory graph linking time-aware gists and facts, achieving 3.4\% and 13.4\% absolute improvements on episodic recollection and reasoning tasks over Mem0 and HippoRAG 2 \cite{shu2025remem}. A-Mem (NeurIPS 2025) further proposed dynamic Zettelkasten-based memory graphs with agentic indexing \cite{lee2025mem}.

Each family offers partial solutions but leaves important gaps around contradiction handling, temporal change, and auditable updates, which NeuSymMS directly targets.

Decades of research in knowledge representation, rule-based reasoning, and truth maintenance systems produced mature symbolic techniques, yet these methods are rarely applied in modern LLM agentic memory~\cite{giarratano2005expert,doyle1979tms}. We contend that high-quality agent memory requires both neural and symbolic components: neural models excel at interpreting language, while symbolic systems offer deterministic reasoning, explicit policies, and auditability over structured facts~\cite{garcez2019neurosymbolic,hogan2021knowledge}.

One of the widely used expert systems is CLIPS, a public-domain expert system tool originally developed at NASA's Johnson Space Center to support the development and deployment of rule- and object-based knowledge systems~\cite{clips_manual}. It provides an integrated environment for representing knowledge via forward-chaining production rules, procedural functions, and an object-oriented layer that supports classes, message-handlers, inheritance, and polymorphism~\cite{clips_manual}. 

In this publication we introduce a production-ready memory subsystem for LLM agents that embodies this hybrid view. We use LLMs to extract atomic facts from dialogue and CLIPS-based expert system to classify, resolve contradictions, and manage memory lifecycles under explicit rules.

The contributions of this publication are:
\begin{itemize}
\item a practical architecture that integrates neuro-symbolic memory for multi-agent platforms;
\item a dual-horizon, access-based memory model inspired by cognitive psychology; and
\item a three-tier scoping that is LLM provider-agnostic and can be plugged with any argentic system
\end{itemize}

\section{Related Work}

Most agentic ``memory'' implementations treat dialogue history as an unstructured log and rely on embedding-based retrieval to select snippets for each turn. Lewis et al.\ investigated retrieval-augmented generation and showed that coupling a parametric language model with a non-parametric document store substantially improves factual recall on knowledge-intensive tasks, but at the cost of additional retrieval latency and pipeline complexity~\cite{lewis2020retrieval}. Karpukhin et al.\ introduced Dense Passage Retrieval (DPR) and demonstrated that dense dual-encoder representations can outperform traditional sparse retrieval for open-domain question answering, yet their approach remains agnostic to contradictions and temporal drift in the underlying corpus~\cite{karpukhin2020dense}. Follow-up work by Izacard and Grave further explored architectures that first retrieve passages and then condition generation on these retrieved segments, highlighting a trade-off between retrieval quality, computational cost, and robustness to noisy context~\cite{izacard2021leveraging}. In all of these retrieval-centric systems, transcripts are treated as immutable documents, and there is no explicit notion of user-level belief revision when information changes.

A second line of work compresses history through abstractive summarization instead of storing or retrieving raw logs. Zhang et al.\ proposed PEGASUS, a pre-training objective based on gap-sentence generation, and showed that abstractive summarization models can produce high-quality compressed representations of long documents~\cite{zhang2020pegasus}. Raffel et al.\ introduced the T5 text-to-text framework and demonstrated that large sequence-to-sequence models can be adapted to a wide variety of summarization and transformation tasks within a unified architecture~\cite{raffel2020exploring}. Laban et al.\ revisited summarization evaluation and found that automatic metrics do not always correlate well with human judgments, especially when summaries omit fine-grained factual details~\cite{laban2021summeval}. Applied to agent memory, these summarization approaches provide concise context but sacrifice atomicity, making it difficult to update or retract individual facts when the user’s state changes.

Recent work has also examined how large language models actually use long contexts. Liu et al.\ analyzed transformer-based models on tasks requiring identification of relevant information in extended input sequences and found that performance often degrades when crucial information appears in the middle of the context window, a phenomenon they term ``lost in the middle''~\cite{liu2023lost}. Beltagy et al.\ developed Longformer, an architecture combining local and global attention patterns to scale transformer models to longer documents while controlling computational cost~\cite{beltagy2020longformer}. These findings suggest that naively appending ever-growing histories even when architectures are modified for long-context processing does not automatically yield reliable use of past information, reinforcing the need for structured, selective memory rather than raw log replay.

Beyond unstructured text, several research studies formalized memory as structured knowledge. Hogan et al.\ surveyed knowledge graphs and showed how triple-based structures support querying, integration, and reasoning ~\cite{hogan2021knowledge}. Nickel et al.\ reviewed relational methods for knowledge graphs and demonstrated that embedding-based models can capture rich relational patterns while still relying on explicitly structured schemas~\cite{nickel2016review}. These works motivate representing agent memory as subject-relation-value triples, but they do not address how to automatically extract, classify, and maintain such triples over evolving conversational histories in production systems.

In this publication, we investigated Classical AI research, which provides tools for rule-based reasoning, truth maintenance, and expert-systems that are highly relevant but underutilized in agentic systems. Forgy introduced the Rete algorithm~\cite{forgy1982rete} and showed that it enables efficient pattern matching for rules, forming the backbone of high-performance rule engines such as CLIPS. Doyle proposed Truth Maintenance Systems (TMS) to maintain logical consistency in dynamic knowledge bases and illustrated how justifications and dependency tracking support belief revision~\cite{doyle1979tms}. De Kleer extended this line of work with assumption-based TMS, demonstrating how explicit assumptions and conflicts can be managed in complex reasoning tasks~\cite{dekleer1986abtms}. Giarratano and Riley’s textbook further systematized design principles for expert systems, emphasizing transparent rule sets and auditable inference chains~\cite{giarratano2005expert}. The CLIPS user’s guide documents a mature, production-ready rule engine originally developed at NASA, provides practical guidance on templates, salience, and working-memory management~\cite{clips_manual}. Despite their strengths for deterministic reasoning and explanation, these symbolic techniques are rarely used as first-class components in contemporary LLM-based agentic platforms.

Hybrid neuro-symbolic approaches aim to combine the pattern recognition capabilities of neural models with the interpretability and logic of symbolic systems. Garcez et al.\ surveyed neural-symbolic learning and reasoning and argued that integrating connectionist models with symbolic knowledge bases can yield systems that both learn from data and support logical inference~\cite{garcez2019neurosymbolic}. Besold et al.\ provided a complementary survey and highlighted a range of architectures that embed symbolic constraints into neural models or use logic as a scaffold for learning~\cite{besold2017neuralsymbolic}. Mao et al.\ introduced the Neuro-Symbolic Concept Learner and showed how visual perception can be combined with symbolic program induction to achieve robust compositional reasoning on visual question answering tasks~\cite{mao2019neurosymbolic}. These works demonstrate the promise of neuro-symbolic integration which can be used for agents' memory in multi-tenant agentic platforms.

Cognitive psychology also offers models that inspire artificial memory lifecycle design. Atkinson and Shiffrin proposed a dual-store theory in which information passes from a limited, short-term buffer to a more durable long-term store through processes such as rehearsal~\cite{atkinson1968human}. Baddeley and Hitch refined this view with the working memory model, showing that short-term retention involves multiple components specialized for different modalities~\cite{baddeley1974working}. Ebbinghaus’s experiments on forgetting revealed that unused information decays over time according to a characteristic curve, underscoring the importance of repeated access for long-term retention~\cite{ebbinghaus1885memory}. NeuSymMS adopts a dual-horizon design and access-based promotion mechanism that are conceptually aligned with these cognitive findings.

Finally, several recent agent architectures explicitly incorporate memory to support more coherent long-term behavior. Park et al.\ introduced generative agents that log experiences, reflect on them, and retrieve relevant memories to plan future actions, demonstrating that such mechanisms can produce believable, persistent behavior in simulated environments~\cite{park2023generative}. Shinn et al.\ proposed Reflexion, where agents store trajectories and verbal feedback and showed that iterative self-reflection over this memory can improve performance on complex tasks~\cite{shinn2023reflexion}. Wang et al.\ presented Voyager, an open-ended embodied agent that accumulates skills and behaviors over time in a persistent world, illustrating how long-term memory enables continual learning~\cite{wang2023voyager}. These systems underscore the value of memory for agent capabilities but typically rely on unstructured logs, code libraries, or task-specific stores rather than a unified, rule-governed knowledge base.

NeuSymMS contributes a production-oriented, hybrid neuro-symbolic memory system that (i) represents user knowledge as scoped triples; (ii) uses LLMs only for neural fact extraction; and (iii) delegates classification, contradiction handling, and lifecycle management to a CLIPS-based expert system. This design bridges retrieval-based and summarization-based approaches by offering a compact, auditable memory layer tailored to multi-agent, multi-tenant agentic platforms.

\section{System Design}

\subsection{Architectural Overview}

NeuSymMS is designed as a memory layer that sits between user conversations and the LLM gateway, providing read and write paths around each agent turn. On the pre-turn read path, the system loads relevant facts for a user, formats them into a compact textual context block, and injects this into the LLM's system prompt via a provider-agnostic gateway. On the post-turn write path, it extracts candidate facts from the interaction, processes them through a CLIPS rules engine, and applies the resulting decisions to a PostgreSQL-backed fact store.

The implementation leverages Django REST Framework for backend APIs, PostgreSQL for durable fact storage, CLIPSPy as Python bindings for CLIPS 6.4, and LiteLLM as an LLM gateway across providers such as OpenAI, Anthropic, and Gemini. Facts are injected as system prompt text, making the memory layer independent of any specific provider's proprietary tools.

\subsection{Triple-Based Data Model}

All knowledge in NeuSymMS is represented as subject--relation--value triples, an abstraction inspired by RDF-style knowledge graphs~\cite{berners2001semantic,hogan2021knowledge}. In this paradigm, each fact is modeled as a directed edge (relation) linking a subject node to an object/value node, enabling graph-structured querying, integration, and reasoning over heterogeneous data at scale. Each triple in NeuSymMS is associated with a user, optional agent and flow (coordinated and interacting set of agents) identifiers, and metadata including a semantic category, a \texttttoken{memory_type} (\texttttoken{short_term} or \texttttoken{long_term}), confidence, \texttttoken{access_count}, and timestamps for provenance and analytics. Soft deletion is implemented via an \texttttoken{is_active} flag to support reversible deactivation and historical analysis.

A simplified schema from the reference implementation is:
\begin{lstlisting}[language=Python]
class MemoryFact(models.Model):
    id = UUIDField(primary_key=True)
    user = ForeignKey(User) # fact owner
    
    # 'user' | 'agent' | 'flow'
    scope = CharField()
    agent_id = IntegerField(null=True)
    flow_id = IntegerField(null=True)
    
    # e.g., "user", "user.pet"
    subject = CharField()
    
    # e.g., "has_pet", "prefers"
    relation = CharField()
    
    # e.g., "golden retriever named Max"
    value = TextField()
    
    # personal, preference, ...
    category = CharField()
    
    # 'short_term' | 'long_term'
    memory_type = CharField()
    
    # 0.0 -- 1.0
    confidence = FloatField()
    access_count = IntegerField()
    source_text = TextField()
    last_accessed_at = DateTimeField()
    is_active = BooleanField()
    
\end{lstlisting}

The system distinguishes nine semantic categories:
\emph{personal}, \emph{preference}, \emph{task}, \emph{relationship}, \emph{skill}, \emph{context}, \emph{instruction}, \emph{temporal}, and \emph{other}. These categories underpin storage policies, promotion rules, and UI filtering---for example, personal and preference facts map naturally to long-term memory, whereas task and temporal facts default to short-term.

\subsection{Three-Tier Scoping}

NeuSymMS enforces a three-tier scoping model to prevent cross-entity contamination: user, agent, and flow scopes. User-scoped facts capture global information about a person, such as name, location, or enduring preferences, while agent-scoped facts describe metrics or state for a particular agent instance. Flow-scoped facts capture workflow-specific context for a coordinated set of agents executing a task together: a flow is a multi-agent workflow in which agents pass messages, follow control flows, and trigger other workflow actions to accomplish a shared objective. Only user-scoped facts are injected into LLM prompts; agent and flow facts are reserved for user selection and analytics, thereby avoiding pollution of a user's identity with data from other entities, such as resumes or third-party documents. 

Automatic scope assignment ensures that when an \texttttoken{agent_id} is provided, extracted facts can be attributed to agent or flow scope as appropriate, with safeguards to avoid misclassification. This scoping design is particularly important in multi-tenant platforms where many agents and flows operate concurrently on behalf of the same user.
\section{Methods}

\subsection{LLM-Based Fact Extraction}

The write path begins by calling a lightweight LLM (e.g., \texttttoken{gpt-4.1-mini} via LiteLLM) with a specialized system prompt that instructs it to act strictly as a fact-extraction engine. Given a conversation snippet consisting of user messages and optional agent responses, the model outputs a JSON array of candidate facts including \texttttoken{subject}, \texttttoken{relation}, \texttttoken{value}, \texttttoken{confidence}, and \texttttoken{scope}. The prompt enforces a low temperature (0.1), a maximum of ten facts per turn, scope tagging, and explicit encoding of negation via relations such as \texttttoken{no_longer_has}, \texttttoken{lost}, or \texttttoken{stopped}....Treating LLMs primarily as extractors (parsers) is a practical design pattern to reduce hallucination and cost compared to using large models for reasoning and retrieval~\cite{yan2023llmpatterns}.

As a concrete example, consider the user utterance:
``I just moved to Toronto and started a new job at iVedh".
The extractor produces a JSON array of atomic facts:
\begin{lstlisting}[language=Python]

{"subject": "user", "relation": "lives_in",
"value": "Toronto", "confidence": 0.95,
"scope": "user"},
{"subject": "user", "relation": "works_at",
"value": "iVedha", "confidence": 0.9,
"scope": "user"}

\end{lstlisting}

This illustrates how free-form text is converted into structured triples with confidence scores and scopes suitable for symbolic processing.

This design treats the LLM as a parser rather than a reasoning engine, aiming for consistency and minimizing hallucination. When extraction models are unavailable, the system degrades gracefully, returning an empty fact list and allowing the platform to operate without memory rather than failing requests.

\subsection{CLIPS Rules Engine}
For each request, candidate facts and existing active facts for a user are passed to CLIPS, preventing working-memory contamination across users or sessions. Three core CLIPS templates---\texttt{memory-fact}, \texttt{candidate-fact}, and \texttt{engine-decision}---encode existing knowledge, incoming candidates, and the decisions to be applied to the database, respectively. Here is how these are defined:

\begin{lstlisting}
(deftemplate memory-fact
   (slot subject (type SYMBOL))
   (slot relation (type SYMBOL))
   (slot value (type STRING))
   (slot confidence (type FLOAT))
   (slot scope (type SYMBOL)))

(deftemplate candidate-fact
   (slot subject (type SYMBOL))
   (slot relation (type SYMBOL))
   (slot value (type STRING))
   (slot confidence (type FLOAT))
   (slot scope (type SYMBOL)))

(deftemplate engine-decision
   (slot fact-id (type INTEGER))
   (slot action (type SYMBOL))
   (slot reason (type STRING)))
\end{lstlisting}

Additional templates capture prompt context metadata such as keywords and intent to support relevance-aware rules:

\begin{lstlisting}
(deftemplate prompt-context
   (slot keywords (type MULTISLOT))
   (slot intent (type SYMBOL))
   (slot turn-number (type INTEGER)))
\end{lstlisting}

Using CLIPS as a production rule engine yields fast forward-chaining over these templates and transparent rule traces that can be inspected for debugging and audit purposes~\cite{forgy1982rete,clips_manual}.
\subsection{Classification Phase}

In the classification phase, CLIPS rules assign a semantic category to every unclassified candidate fact based on pattern matching over \texttttoken{relation} and \texttttoken{value} fields. A cascade of rules maps relations like \texttttoken{has_pet}, \texttttoken{family}, \texttttoken{lives_in}, \texttttoken{prefers}, or \texttttoken{speaks_language} to categories such as personal, preference, or skill, with a catch-all rule that labels remaining facts as other. Implementing classification in CLIPS rather than via an LLM provides determinism, microsecond-level performance, zero API cost, and transparent mapping from patterns to categories \cite{giarratano2005expert}.

\subsection{Contradiction Detection and Truth Maintenance}

The second phase implements truth maintenance, ensuring that the knowledge base does not contain inconsistent facts about the same subject and relation. In the reference implementation, entity normalization combines lower-casing, Unicode normalization, and simple lexical cleanup (e.g., trimming punctuation and whitespace) before comparison. For contradictions we apply approximate string matching (Levenshtein distance over normalized values) with a fixed similarity threshold (0.85 in our deployment) to decide whether two values refer to the same underlying entity (e.g., employer names or locations). For inherently multi-valued relations such as \textit{tokenspeakslanguage}, the engine treats each distinct value as a separate fact and only retracts the specific matched value on negation, while single-valued relations such as \textit{tokenworksat} and {tokenlivesin} default to a latest-value-wins policy. Negations are encoded as relations such as \textit{tokennolongerhas}, \textit{tokenstopped}, or \textit{tokenlost} by the extractor; rules translate these into retractions of positive facts and do not persist the negated form as a long-term fact. One key rule triggers when a candidate fact shares subject and relation with an existing \texttttoken{memory-fact} but has a different value, leading to an \texttttoken{engine-decision} that retracts the old fact and stores the new one, typically as \texttttoken{long_term} to reflect corrected knowledge. These mechanisms mirror classic truth maintenance and belief revision techniques \cite{doyle1979tms,dekleer1986abtms}.

{Negation Example :} Suppose the system stores a fact
\begin{verbatim}
(memory-fact subject="user" relation=
"has_pet" value="cat named Whiskers")
\end{verbatim}
and later observes the utterance:
\begin{quote}
My cat died.
\end{quote}
The extractor emits a negated candidate such as
\begin{verbatim}
(candidate-fact subject="user" relation=
"died" value="cat named Whiskers")
\end{verbatim}
A CLIPS rule detects the negation via the \texttttoken{died} relation and
a normalized string comparison on the value (e.g., lower-casing and
whitespace/punctuation cleanup), emits a decision to retract the positive
\texttttoken{has_pet} fact, and discards the negation candidate rather than
storing it as enduring knowledge.
\subsection{Promotion, Storage, and Pruning}

The final phase determines storage policies and fact lifecycles. High-salience pre-filters discard low-confidence facts (e.g., confidence $< 0.3$) and skip exact duplicates of existing active facts, reducing noise and redundant storage. In our production deployment, we discard candidate facts with confidence below 0.3, initialize \textit{tokenaccesscount} to 0 on write, promote \textit{tokenshortterm} facts to \textit{tokenlongterm} when \textit{tokenaccesscount} $>= 3$, and run a background job that deactivates \textit{tokenshortterm} facts older than 24 hours with \textit{tokenaccesscount} = 0. NeuSymMS currently represents temporal validity using creation and last-access timestamps per fact rather than explicit validity intervals; we defer richer interval reasoning to future work. Auto-promotion rules store personal, preference, instruction, and skill facts directly as \texttttoken{long_term}, reflecting their durability and importance for user experience.

Other categories default to \texttttoken{short_term}, subject to an access-based promotion rule: \texttttoken{short_term} facts whose \texttttoken{access_count} reaches or exceeds a fixed threshold (e.g., three) are promoted to \texttttoken{long_term}. A background pruning process deactivates \texttttoken{short_term} facts older than a configured time-to-live (e.g., 24 hours) that have low \texttttoken{access_count}, whcich infact is a dual-horizon memory model with decay for unused information. This design uses \texttttoken{access_count} as an organic importance signal derived from actual agent usage rather than a separate importance-scoring model, an idea aligned with cognitive theories of rehearsal and consolidation \cite{atkinson1968human,baddeley1974working,ebbinghaus1885memory}.

\subsection{End-to-End Write Path Example}

To illustrate the full write path, consider the following interaction:

``I just started a new job at Google in Mountain View. I speak Python and Go."

\paragraph{Step 1: Extraction}
The extractor yields:
\begin{verbatim}
[
{subject: "user", relation: "works_at", 
value: "Google", confidence: 0.95},
{subject: "user", relation: "lives_in", 
value: "Mountain View", confidence: 0.85},
{subject: "user", relation: 
"speaks_language", 
value: "Python", confidence: 0.9},
{subject: "user", relation: 
"speaks_language", 
value: "Go", confidence: 0.9}
]
\end{verbatim}

\paragraph{Step 2: Existing Facts}
Suppose the database currently contains:
\begin{verbatim}
fact_id: "abc", subject: "user", relation:
"works_at", value: "Meta"
fact_id: "def", subject: "user", relation:
"lives_in", value: "Menlo Park"
\end{verbatim}

\paragraph{Step 3: Classification}
Relations such as \texttttoken{works_at} and \texttttoken{lives_in} are classified
as \emph{personal}, while \texttttoken{speaks_language} is classified as a
\emph{skill} category.

\paragraph{Step 4: Contradiction Detection}
The candidates \texttttoken{works_at Google} and \texttttoken{lives_in Mountain View}
conflict with existing facts \texttttoken{works_at Meta} and
\texttttoken{lives_in Menlo Park}. The engine therefore emits decisions to
retract the old facts and store the new ones as long-term knowledge, demonstrating belief revision in practice \cite{doyle1979tms}.

\paragraph{Step 5: Promotion and Storage}
Skill facts such as \texttttoken{speaks_language Python} and
\texttttoken{speaks_language Go} are auto-promoted to long-term memory
by category-based rules. The resulting database state includes the
updated employer and location, plus two durable language skills.

\subsection{Read Path and Context Injection}

On the read path, NeuSymMS queries \texttttoken{NeuSymMS_memory_facts} for active, user-scoped facts for the given user, ordering first by \texttttoken{memory_type} (long-term before short-term), then by \texttttoken{access_count}, and finally by recency. The system caps the number of injected facts (e.g., 30) to avoid overloading prompts and formats them into a text block headed by ``[Memory -- Known facts about this user]'' with simple labels for memory type and category. This read-inject pattern differs from retrieval/embedding pipelines like RAG~\cite{lewis2020retrieval} and DPR~\cite{karpukhin2020dense}, which inject large context windows or embedding-selected snippets. NeuSymMS instead directly injects concise structured facts, avoiding overloading prompts and maintaining atomicity.

\subsection{Read Path Example}

As an example of the read path, assume the database contains long-term
facts:
\begin{verbatim}
user works_at Google
user lives_in Mountain View
user speaks_language Python
user speaks_language Go
user prefers dark mode
\end{verbatim}

When the user asks:
``What languages do I know?"
the system issues a query for active user-scoped facts, orders them by
memory type, access count, and recency, and formats a context block:
\begin{verbatim}
[Memory -- Known facts about this user]
[LT/personal] user works_at Google
[LT/personal] user lives_in Mountain View
[LT/skill] user speaks_language Python
[LT/skill] user speaks_language Go
[LT/preference] user prefers dark mode
\end{verbatim}
This block is injected into the LLM's system prompt, enabling responses
such as: ``Based on what I know about you, you're proficient in Python
and Go.''

\section{Implementation}

\subsection{Technology Stack}

NeuSymMS is implemented as a Memory System inside the Nexa platform~\cite{nexa_platform}, exposed both via backend APIs and a UI. The backend uses Django and Django REST Framework to manage a \textit{tokenNeuSymMSmemoryfacts} model and related endpoints. CLIPSPy provides Python bindings for CLIPS 6.4, enabling the rules engine to run in-process with low latency and no external dependencies \cite{clips_manual,forgy1982rete}. LiteLLM serves as a provider-agnostic gateway for LLM calls, abstracting over OpenAI, Anthropic, Google, Azure, and other backends.

The frontend exposes a Memory tab in the Settings UI built with React and TypeScript, allowing users to view, search, filter, edit, and clear memory facts. Backend integration with agents and flows is achieved via utility functions such as \texttttoken{enrich_instructions_with_memory} and \texttttoken{process_memory}, which are called from agent and flow executors to ensure that every LLM interaction can both consume and update memory.

\subsection{Database Schema and Indexes}

The \texttttoken{NeuSymMS_memory_facts} table includes several targeted indexes: \texttttoken{(user, is_active, scope)} for read-path queries of user-scoped active facts; \texttttoken{(user, agent_id, is_active)} and \texttttoken{(user, flow_id, is_active)} for agent- and flow-scoped queries; \texttttoken{(user, category, is_active)} for UI filtering; and \texttttoken{(subject, relation)} to support contradiction detection by efficiently finding existing facts sharing a subject--relation pair. These indexes enable efficient access patterns for core operations while controlling storage overhead.

\subsection{API Surface and UI}

A Django REST API provides endpoints to list, summarize, update, delete, and clear memory facts, enabling both programmatic access and integration with external tools. Because NeuSymMS is integrated into the Nexa platform, memory operations participate in the same user and access-control model as documents and agents: only authenticated users can view or edit their own facts; tenant administrators can configure classification and retention policies; and all edits, including Clear All, are served via audited REST endpoints consistent with the platform's user management and security model. The Memory tab UI shows summary cards for active, long-term, and short-term facts; category and type filters; a searchable fact list with inline editing; and a bulk ``Clear All" operation with optional scoping. This human-in-the-loop interface provides transparency into what the system knows and supports user control for privacy and compliance.

\section{Results and Discussion}

\subsection{Salient Features and Novelty}

NeuSymMS's most distinctive feature is its hybrid neuro-symbolic architecture: LLMs are used only where they add value (natural language understanding), while CLIPS handles deterministic tasks such as classification, contradiction detection, and lifecycle management. This separation yields zero-cost, low-latency decisions with explicit rule-based justification via \texttttoken{reason} fields in \texttttoken{engine-decision} facts.

The system implements automated truth maintenance, retracting stale facts when new information arrives, and negation-aware extraction that properly handles statements about loss, cessation, or change. Its access-based promotion mechanism provides an organic importance signal without additional models, and three-tier scope isolation prevents cross-entity contamination in multi-agent scenarios. Human-in-the-loop management through the UI further enhances trust and regulatory compliance by making memory contents visible and editable.

\subsection{Reliability and Cost Considerations}

NeuSymMS's design prioritizes graceful degradation. If LLM extraction fails, memory simply does not update; if the CLIPS engine encounters an error, the system can fall back to storing candidates as short-term; if memory loading fails, agents continue without context rather than failing entire requests. These design choices ensure that memory remains an enhancement rather than a single point of failure.

Cost is minimized by using a low cost LLM, running CLIPS locally, and avoiding embedding-based retrieval or additional scoring models. Context injection is implemented as string concatenation of selected facts, eliminating the need for expensive vector search at inference time. This makes the system suitable for high-throughput where per-turn costs must remain tightly bounded.

\section{Conclusion and Future Work}

We have presented NeuSymMS, a practical neuro-symbolic memory system for LLM agents that combines neural fact extraction with a CLIPS-based expert system for deterministic, auditable memory management. By representing knowledge as triples, enforcing scoped lifecycles, and leveraging access-based promotion, NeuSymMS provides agents with durable, evolving knowledge of their users while controlling cost and ensuring reliability.

Future work includes richer temporal reasoning over intervals and events, improved negation and uncertainty handling, and automated rule-learning or rule-suggestion mechanisms that could adapt classification and lifecycle policies to new domains. We also plan to evaluate user experience and downstream task performance with and without NeuSymMS, to quantify the impact of structured memory on real-world agent deployments.

\subsection{Evaluation Roadmap}
A full quantitative evaluation is left to future work, but we outline our planned protocol to situate NeuSymMS among recent structured memory systems. We plan to evaluate on long-horizon memory benchmarks such as LoCoMo~\cite{locomo_acl24}, LongMemEval~\cite{longmemeval_wu24}, AMA-Bench~\cite{ama_bench_2602}, and MEMORYARENA~\cite{memoryarena_website}, using fixed LLM backbones to isolate the effect of the memory layer. Baselines may include Mem0~\cite{mem0_2504}, MemGPT/AMEM~\cite{memgpt_2310}, LiCoMemory~\cite{licomemory_2511}, TELEMEM~\cite{telemmem_2601}, GraphRAG/Semantic Anchoring~\cite{semantic_anchoring_2508}, and AMAC~\cite{amac_2603}, allowing head-to-head comparisons of contradiction handling, retrieval quality, and operational cost.

We will report agent-centric metrics such as task success, retrieval accuracy, and multi-session consistency, alongside system metrics including latency ($T_\texttttoken{read}$, $T_\texttttoken{write}$), token and compute cost, and throughput under multi-tenant load. 

\bibliographystyle{IEEEtran}
\bibliography{references}

\end{document}